%% file: root.tex
\pdfoutput=1
\documentclass[conference,final]{IEEEtran}
\IEEEoverridecommandlockouts
\usepackage[utf8]{inputenc}
\usepackage[T1]{fontenc}
\usepackage[american]{babel}
\usepackage{csquotes}

\usepackage{graphicx}
\graphicspath{{./figures/}}

\usepackage{mathtools}
\usepackage{amsmath}
\usepackage{amssymb}
\usepackage{gensymb} %
\usepackage{textcomp}

\usepackage[dvipsnames]{xcolor}
\usepackage{customcolors}

\usepackage{eccvabbrv}
\usepackage{import}
\usepackage{tabularray}
\UseTblrLibrary{booktabs}
\UseTblrLibrary{diagbox}

\usepackage[ruled,vlined,linesnumbered]{algorithm2e}
\SetKwInput{KwInput}{Input}
\SetKwInput{KwOutput}{Output}

\usepackage{cite}
\usepackage[bookmarks=true,colorlinks=true,linkcolor=black,citecolor=black,urlcolor=black]{hyperref}

\usepackage{cleveref}
\crefname{figure}{Fig.}{Figs.}
\Crefname{figure}{Fig.}{Figs.} %

\begin{document}
\def\IEEEtitletopspaceextra{0.25in}
\title{REST: Receding Horizon Explorative Steiner Tree\\for Zero-Shot Object-Goal Navigation}
\author{
  \IEEEauthorblockN{Shuqi Xiao\textsuperscript{1},
    Maani Ghaffari\textsuperscript{2},
    Chengzhong Xu\textsuperscript{1},
    and Hui Kong\textsuperscript{1*}
    \thanks{\textsuperscript{*}\,Corresponding author: Hui Kong.}
    \thanks{This work was supported by the Fundo para o Desenvolvimento das Ci\^encias e da Tecnologia of Macau (FDCT) under Grant No.~0067/2023/AFJ and No.~0117/2024/RIB2.}
  }
  \IEEEauthorblockA{\textsuperscript{1}State Key Laboratory of Internet of Things for Smart City (SKL-IOTSC), University of Macau}
  \IEEEauthorblockA{\textsuperscript{2}Department of Naval Architecture and Marine Engineering, and Department of Robotics, University of Michigan}
  \IEEEauthorblockA{\texttt{shuqi.xiao@connect.um.edu.mo}, \texttt{maanigj@umich.edu}, \texttt{czxu@um.edu.mo}, \texttt{huikong@um.edu.mo}}}
\maketitle
\bstctlcite{BSTcontrol}
\begin{figure*}[t]
  \centering
  \includegraphics[width=0.9\linewidth]{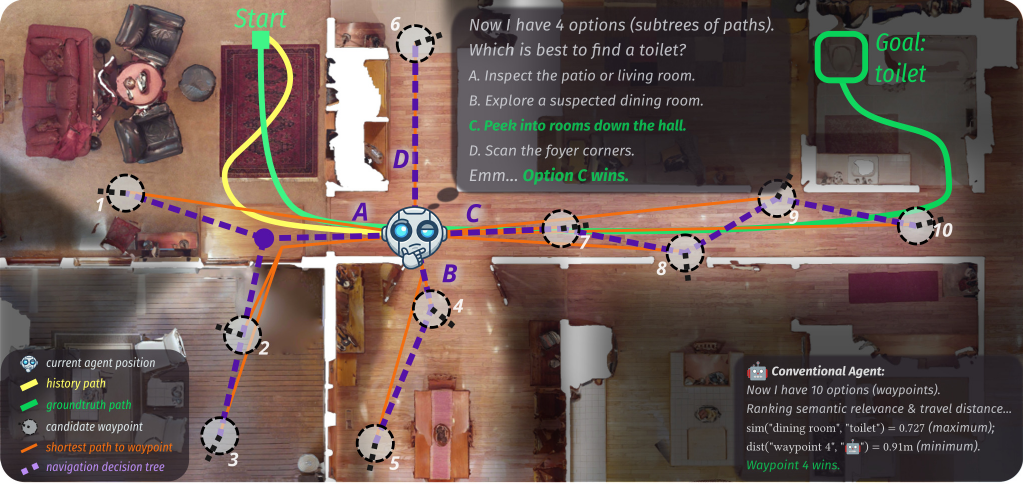}
  \caption{
    REST reasons over an agent-centric tree of safe and informative
    paths (purple) rather than evaluating isolated candidate
    waypoints (gray) like conventional hierarchical ObjectNav agents. Here, REST
    selects the next-best subtree among 4 options via spatial
    narratives; a conventional agent (\eg,
    VLFM~\cite{yokoyamaVLFMVisionLanguageFrontier2024}) independently
    scores 10 waypoints by semantic similarity and geometric
    proximity, discarding spatial-temporal context.
  }
  \label{fig:teaser}
\end{figure*}
\input{abstract}
\section{Introduction}\label{sec:intro}
\input{introduction/a}
\input{introduction/b}
\input{introduction/c}
\section{Related Work}
\label{sec:related-work}
\input{related-work/b}

\input{related-work/c}
\section{Methodology}
\label{sec:methodology}
\input{methodology/methodology}

\section{Experiments}
\label{sec:experiments}
\input{experiments/experiments}

\section{Conclusion}
\input{conclusion/conclusion}
\bibliographystyle{IEEEtran}
\bibliography{IEEEabrv,confabrv_short,references}
\end{document}

%% file: abstract.tex
\begin{abstract}
  Zero-shot object-goal navigation (ZSON) requires navigating unknown
  environments to find a target object without task-specific training.
  Prior hierarchical solutions mainly focus on either scene understanding and
  representations (\textbf{\textit{belief}}) or high-level
  decision-making and planning (\textbf{\textit{policy}}), yet treat
  the \textit{option}, \ie, the subgoal candidate that belief
  proposes and policy selects, as an interface
  inherited from adjacent modules rather than a design axis in its own right.
  In practice, options are predominantly single waypoints scored by
  destination utility: a lone destination hides the value gathered en
  route, and a flat list obscures the relationships among candidates.
  Our insight is that the option space should be a \textit{tree of
  paths}. Full paths expose en-route information gain that
  destination-only scoring systematically neglects; a tree of shared
  segments enables coarse-to-fine LLM reasoning that dismisses or
  pursues entire branches before examining individual leaves,
  compressing the combinatorial path space into an efficient hierarchy.
  We instantiate this insight in \textbf{REST} (Receding Horizon
  Explorative Steiner Tree), a training-free framework that (1)
  builds an explicit open-vocabulary 3D map from online RGB-D
  streams; (2) grows an agent-centric tree of safe and informative
  paths as the option space via sampling-based planning; and (3)
  textualizes each branch into a spatial narrative and selects the
  next-best path through chain-of-thought LLM reasoning.
  Across the Gibson, HM3D, and HSSD benchmarks, REST consistently
  ranks among the top methods in success rate and path efficiency.
\end{abstract}

%% file: introduction/a.tex
\par Object-goal Navigation (ObjectNav) requires an embodied agent to navigate to an object of a specified category (\eg, coffee cup) within a previously unseen environment~\cite{batraObjectNavRevisitedEvaluation2020}.
Whereas a Vision-Language Navigation (VLN) agent is guided by step-by-step linguistic instructions, an ObjectNav agent must rely on itself to reason and plan strategies that balance exploring unknown spaces and exploiting observed cues toward the goal.
To acquire such autonomy from experience, fueled by high-fidelity simulators and massive 3D datasets, deep reinforcement and imitation learning have driven significant progress~\cite{duVTNetVisualTransformer2021,ramrakhyaPirlnavPretrainingImitation2023,chenObjectGoalNavigation2023}.
However, these end-to-end ObjectNav policies remain brittle under distribution shifts (\eg, closed vocabulary, sim-to-real gap) and demand large volumes of high-quality interaction data, limiting their generalizability and scalability.
In recent years, the zero-shot capabilities of AI foundation models, \eg, Vision-Language Models (VLM) and Large Language Models (LLM), have catalyzed a paradigm shift toward zero-shot object-goal navigation (ZSON), in which no task-specific training is required.
Since Majumdar~\etal~\cite{majumdarZSONZeroshotObjectgoal2022} first demonstrated this concept by transferring an image-goal navigation policy to object-goal tasks via CLIP~\cite{radfordLearningTransferableVisual2021},
subsequent work is increasingly converging on fully training-free and hierarchical architectures~\cite{zhouESCExplorationSoft2023,gadreCoWsPastureBaselines2023,yinSGNavOnline3D2024,wuVoroNavVoronoibasedZeroshot2024} that marry the two paradigms: data-driven foundation models supply generalizable semantic understanding across novel objects and environments, while model-based navigation frameworks, \eg, SLAM and path planning, provide safety and efficiency guarantees grounded in geometric principles.

\par These hierarchical ObjectNav agents can be generally characterized by three axes:
(1) \textit{belief update}: provides prior commonsense knowledge (\eg, VLM) and aggregates posterior observations (\eg, mapping);
(2) \textit{option space}: represents the set of candidate subgoals;
(3) \textit{hierarchical policy}: a global policy selects the next-best option and a local policy maps a high-level option to low-level actions.
Prior work has invested heavily in enriching the belief, \eg, semantic value map~\cite{yokoyamaVLFMVisionLanguageFrontier2024,fangGAMapZeroShotObject2024} or open-vocabulary 3D scene graphs~\cite{yinSGNavOnline3D2024}, and strengthening the global policy, \eg, LLM-based ranking~\cite{yuL3MVNLeveragingLarge2023,yinSGNavOnline3D2024} or heuristic-based scoring~\cite{wuVoroNavVoronoibasedZeroshot2024,zhouESCExplorationSoft2023}.

%% file: introduction/b.tex
However, the option space, which is the interface that mediates information flow from belief to policy, has drawn comparatively little design attention, more often inherited from whatever representation adjacent modules produce or consume than shaped to serve the policy's reasoning.
Prior hierarchical ObjectNav agents adopt diverse option representations: pixel coordinates predicted by VLMs~\cite{caiBridgingZeroshotObject2024,jinPanoNavMaplessZeroShot2025}, metric map coordinates predicted by RL-based policies~\cite{chaplotObjectGoalNavigation2020,georgakisLearningMapActive2022}, exploration frontiers at the boundary of known and unknown space~\cite{yokoyamaVLFMVisionLanguageFrontier2024,yinSGNavOnline3D2024,zhangApexNAVAdaptiveExploration2025,zhouESCExplorationSoft2023,yuL3MVNLeveragingLarge2023}, and topological graph nodes that simplify the navigable space~\cite{wuVoroNavVoronoibasedZeroshot2024}.
Despite this variety, these representations share a common trait: each candidate is scored by the utility of its destination rather than the path taken to reach it, and in most cases is evaluated independently of the others.
We term this the \textit{next-best-waypoint} paradigm: at each decision epoch, the agent scores candidates by the arrival utility at a waypoint, blind to what lies en route and, in most formulations, to the structural relationships among candidates.
\par Consider the search for a target like a \textit{toilet}, as illustrated in \cref{fig:teaser}.
A waypoint-based agent might deprioritize the end of a long hallway because the destination itself lacks semantic cues and incurs high travel cost.
However, traversing that corridor offers the chance to ``peek'' into multiple passing doorways, one of which likely contains the goal.
An effective option space should therefore be \textit{path-grounded}, making the cumulative gain and cost of the trajectory explicit to the policy.
Transitioning from points to paths, however, introduces a combinatorial explosion: $n$ reachable waypoints can yield an exponential number of possible routes, far exceeding the reasoning capacity of LLM-based policies.
Furthermore, listing all candidates independently obscures their shared intent: in \cref{fig:teaser}, the paths to waypoints 7--10 all traverse the corridor, yet appear as four separate options, forcing the policy to evaluate route-level variations when the underlying decision, which region to explore next, is the same.
We propose that both challenges are addressed by a \textit{navigation decision tree} that hierarchically clusters candidate paths by identifying their shared segments.
Such a structure enables coarse-to-fine reasoning under partial observation, mirroring human navigation where one first selects a promising direction before refining the specific destination based on new observations.
As shown in \cref{fig:teaser}, dozens of potential paths are compressed into four compact subtrees, allowing the LLM to select the optimal ``branch'' through efficient linguistic reasoning.
We therefore argue that the option space for zero-shot ObjectNav should be a \textit{tree of paths}: path-grounded to expose en-route utility, and tree-structured to enable tractable, hierarchical reasoning.

%% file: introduction/c.tex
\par We instantiate this insight in REST (Receding Horizon
Explorative Steiner Tree), a training-free ObjectNav framework.
REST replaces the conventional set of independently scored waypoints with a
compact \textit{navigation decision tree} that organizes
path-grounded options into an efficient hierarchy.
We evaluate REST on the Gibson, HM3D, and HSSD ObjectNav benchmarks.
Across all three benchmarks, REST consistently ranks among the top
methods in success rate while achieving the best or second-best path
efficiency, demonstrating a favorable efficiency-success balance that
we attribute to tree-structured, path-grounded planning.
We make three contributions:
\begin{itemize}
  \item We recast the option space for zero-shot ObjectNav from
    destination-scored waypoints to the \textit{next-best-path}
    formulation, a path-grounded representation that exposes en-route
    information gain.
  \item We propose the \textit{navigation decision tree}, a
    hierarchical option representation constructed via Euclidean
    Steiner Tree optimization and a LLM-based open-vocabulary
    textualization and decision-making pipeline.
  \item We present REST, a training-free ZSON framework, and
    demonstrate consistently competitive performance across three
    ObjectNav benchmarks.
\end{itemize}

%% file: related-work/b.tex
\subsection{Foundation models for ObjectNav}
\par Foundation models provide transferable priors that reduce or remove the need for task-specific training in embodied agents.
A prominent line of work~\cite{yokoyamaVLFMVisionLanguageFrontier2024,fangGAMapZeroShotObject2024,zhangApexNAVAdaptiveExploration2025} uses CLIP-style VLMs~\cite{radfordLearningTransferableVisual2021} for both open-vocabulary perception and high-level decision-making. These methods project goal-conditioned cosine similarities onto a top-down grid map, producing an implicit ``value map'' that guides frontier selection.
However, cosine similarities among CLIP image or text embeddings can only measure semantic relevance rather than performing complex linguistic reasoning. In the example in \cref{fig:teaser}, \(\langle \text{toilet}, \text{dining room} \rangle \approx 0.73 > 0.67 \approx \langle \text{toilet}, \text{corridor} \rangle\) leads the agent to search the dining room, unaware of the benefits of traversing the corridor to locate a bathroom.
Moreover, since every cell of the grid map encodes relevance to a single target category, the entire map must be discarded and recomputed whenever the goal switches, preventing reuse of previously gathered spatial knowledge.
These two limitations motivate the separation in our framework between a goal-agnostic open-vocabulary semantic map, which can persist across goal changes, and a language-model reasoning layer that exploits commonsense priors over the map.

%% file: related-work/c.tex
\subsection{Autonomous exploration for ObjectNav}
\par In ObjectNav, exploration is inseparable from goal seeking: the
agent must decide where to look next so that each observation gathers
semantic cues that guide it toward the target.
Map-based ObjectNav systems therefore borrow epistemic heuristics
from autonomous exploration to reduce uncertainty about the unknown
environment efficiently.
Frontier-based
exploration~\cite{yamauchiFrontierbasedApproachAutonomous1997} has
been widely
used~\cite{yokoyamaVLFMVisionLanguageFrontier2024,yinSGNavOnline3D2024,zhangApexNAVAdaptiveExploration2025,zhouESCExplorationSoft2023,yuL3MVNLeveragingLarge2023},
and VoroNav~\cite{wuVoroNavVoronoibasedZeroshot2024} builds a
Generalized Voronoi Graph
following~\cite{jiyeongleeSensorbasedExplorationConvex2005} to
provide a graph of collision-free paths for geometric exploration.
Frontier-based methods follow an ``explore-then-plan''
paradigm~\cite{lindqvistTreeBasedNextBestTrajectoryMethod2024}: they
first select a next-best frontier, then invoke a path planner such as
A*~\cite{hartFormalBasisHeuristic1968} to reach it.
Because each option in our framework is a complete, executable path,
exploration and path planning are necessarily joint rather than sequential.
Among prior ZSON methods, VoroNav is the most similar to REST: both
structure the option space as a topological graph of paths rather
than a set of independently scored waypoints.
The distinction lies in how that topology is formed.
VoroNav's graph is a Voronoi skeleton that maximizes clearance from
obstacles, so its topology is dictated entirely by the geometry of
the \emph{known} free space.
This binds the agent's options to a single fixed objective, namely
staying maximally clear of walls, and cannot express the other
intents that ObjectNav demands.
An agent confined to maximal-clearance paths, for instance, cannot
approach a table closely enough to inspect whether a small target rests on it.
More fundamentally, since the target location is unknown, the paths
an agent needs are shaped by what it has yet to observe rather than
by where walls lie, so a topology fixed by known geometry cannot
track an evolving belief state.
REST instead \emph{decouples} the decision topology from the known
geometry: building on sampling-based informative path
planning~\cite{schmidEfficientSamplingBasedMethod2020,lindqvistTreeBasedNextBestTrajectoryMethod2024},
it grows a tree whose terminals are placed by an intent-driven
sampler and then compacted by a Steiner-tree solver.
Because the Steiner solver consumes only terminal points, the policy
that places them becomes a pluggable module: we currently instantiate
it with a geometric information-gain sampler, but learned or
heuristic sub-policies, each encoding a distinct embodied intent such
as close-up inspection, can be added as further terminals without
changing the representation, an extensibility that a fixed Voronoi
skeleton cannot offer.

%% file: methodology/methodology.tex
\par
\begin{figure*}[htbp]
	\centering
	\includegraphics[width=0.95\linewidth]{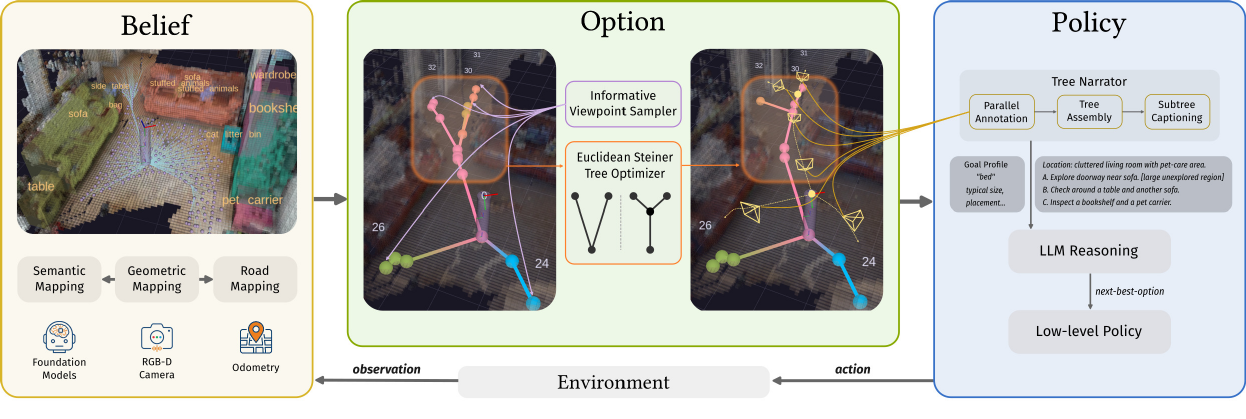}
	\caption{Overview of REST, a training-free ObjectNav framework that replans in a receding-horizon manner. At each decision cycle, the agent updates the from online RGB-D streams, grows an agent-centric Steiner tree of safe and informative paths as the option space, textualizes each branch into a spatial narrative, and selects the next-best path through chain-of-thought LLM reasoning.
	}
	\label{fig:framework}
\end{figure*}

 An overview of REST is shown in Fig.~\ref{fig:framework}.
The system is organized around three components: \textit{belief}, \textit{option}, and \textit{policy}.
At each decision epoch, the agent first updates its belief through geometric, semantic, and road mapping (Sec.~\ref{sec:belief}).
The option space is then constructed by the Receding Horizon Explorative Steiner Tree (Sec.~\ref{sec:option}), which maintains a navigation decision tree of safe, informative, and hierarchical paths; each branch is textualized by projecting the semantic map along a virtual panoramic camera sweep.
Finally, the policy (Sec.~\ref{sec:policy}) selects the next-best-path via LLM reasoning over the textualized options and executes it with an information-gain-driven orientation policy that determines camera headings along the selected path.

\subsection{Belief Update}
\label{sec:belief}
\input{methodology/belief/geometric-mapping}
\input{methodology/belief/semantic-mapping}
\input{methodology/belief/road-mapping}

\subsection{Option Space}
\label{sec:option}
\input{methodology/option/a}

\input{methodology/option/b}

\subsection{Hierarchical Policy}
\label{sec:policy}
\input{methodology/policy/a}
\input{methodology/policy/b}

%% file: methodology/belief/geometric-mapping.tex
\subsubsection{Geometric mapping}
We leverage UFOMap \cite{dubergUFOMapEfficientProbabilistic2020}, an efficient 3D volumetric mapping system, to integrate a posed RGB-D stream into an octree of occupancy voxels online. Each octree node is categorized as unknown, free, or occupied based on its occupancy value.
The geometric map mainly serves as an anchor for semantic mapping and road mapping, and also provides efficient geometry intersection computation
for both collision-checking in safe path planning and ray-casting for informative viewpoint sampling.

%% file: methodology/belief/semantic-mapping.tex
\subsubsection{Semantic mapping}~\label{subsubsection:semantic-mapping}
\paragraph{2D perception}
\par We capture open-vocabulary, explicit and fine-grained 2D semantics by a cascaded \textit{recognize-detect-segment} workflow based on off-the-shelf edge-friendly foundation models.
For each RGB image, first, we prompt a Small Vision-Language Model (SVLM) to tag the image by an array of salient, discrete, manipulable entities with clear affordances (\eg, washing machine, door, plant) it contains.
Then, we feed the entity labels, \eg, wooden cabinet, into an Open-Vocabulary Object Detector (OVOD) to obtain 2D bounding boxes with confidence scores.
Finally, we use an edge-optimized universal segmentation model based on SAM2~\cite{ravi2024sam2} to refine detection boxes to segmentation masks.
This pipeline synergizes these foundation models to overcome their individual limitations on balancing semantic understanding capabilities and visual grounding granularity.
While VLMs offer strong zero-shot understanding, they lack localization precision; conversely, OVOD and SAM excel at dense grounding and segmentation but require explicit text or spatial prompts.
The final results are consumed by the 3D semantic mapping described next.

\paragraph{3D mapping}
First, following~\cite{mccormacSemanticFusionDense3D2017}, we annotate the 3D volumetric map by fusing 2D perception results across multiple viewpoints.
For each frame, we retrieve all visible occupied octree voxels inside the camera's view frustum and project them onto the image plane.
Each projected pixel that falls inside a detected instance mask casts a vote, weighted by detection confidence, for the corresponding instance label.
Each voxel maintains a per-label vote histogram; the label with the highest accumulated score serves as the current semantic estimate.
By aggregating confidence-weighted votes across multiple viewpoints, this majority-voting scheme smooths out single-frame noise and yields a robust 3D semantic map.
Second, we periodically apply per-label DBSCAN~\cite{esterDensitybasedAlgorithmDiscovering1996} to suppress floating artifacts and fit a 3D bounding box to each surviving cluster, producing a set of discrete entities from the annotated voxel map.
To track entities over time, we match clusters between consecutive snapshots by computing 3D IoU of their bounding boxes; a match above a threshold carries forward the existing entity identity, while unmatched clusters are initialized as new entities.
However, open-vocabulary tagging introduces cross-frame label inconsistency: the same real-world cup may be tagged as ``blue cup'', ``ceramic cup'', or ``teacup'' across different frames, producing redundant overlapping entities.
We resolve this with a spatial-semantic joint clustering: we measure spatial overlap by intersection over smaller (IoS) and semantic relevance by the cosine similarity between label embeddings (MobileCLIP2~\cite{faghriMobileCLIP2ImprovingMultimodal2025}). When both exceed their thresholds (\(\tau_{IoS}, \tau_{sim}\)), the smaller entity is merged into the larger one, consolidating fragmented labels into a single block that represents one real-world entity.

%% file: methodology/belief/road-mapping.tex
\subsubsection{Road Mapping}
\par Our path-grounded option space (\cref{sec:option}) requires a library of all feasible paths. Inspired by sampling-based path planning methods,
we adopt Real-Time RRT* (RT-RRT*)~\cite{naderiRTRRTRealtimePath2015} to continuously maintain a global tree of feasible paths rooted at the agent's pose.

We expand the RT-RRT* tree by sampling candidate waypoints with the hybrid local-global strategy of Schmid~\etal~\cite{schmidEfficientSamplingBasedMethod2020}: when node density in the agent's vicinity falls below a threshold, samples are drawn locally to maintain connectivity; otherwise, samples are drawn globally to extend coverage.
Each candidate must pass a traversability check.
We attach a thin axis-aligned bounding box (AABB) beneath the agent's base and query the occupied voxels within this slab on the 3D occupancy grid; the waypoint is accepted only if the occupied-volume ratio exceeds a support threshold.
This check rejects waypoints over open air, which in multi-storey scenes prevents a wheeled robot from falling downstairs.
For each candidate edge, we approximate the robot body with a cylindrical bounding volume, compute the 3D swept volume of the straight-line motion via an oriented bounding box (OBB) estimated by PCA, and declare the edge feasible only if this volume has no intersection with occupied or unknown voxels.
Together, these predicates ensure that every node is traversable and reachable by collision-free edges.

The tree is expanded following the RRT* rewiring procedure~\cite{naderiRTRRTRealtimePath2015}: the new node is connected to its lowest-cost neighbor and nearby edges are rewired if a shorter path through the new node exists.
This rewiring rule is the standard mechanism that gives RRT* its asymptotic path-length optimality.
As the agent moves, the tree is rerooted to the agent's current pose, keeping path queries agent-centric without rebuilding the tree from scratch.
The tree persists across decision epochs and grows incrementally with each map update, eventually spanning all traversable and reachable space observed so far.
The resulting roadmap records where the agent can stand and how it can get there at lowest cost.
The option space (\cref{sec:option}) queries this roadmap to extract informative subgoals and to compute the navigation decision tree.

%% file: methodology/option/a.tex
\subsubsection{Informative viewpoint sampler}
Our RT-RRT* roadmap provides probabilistically complete coverage of
reachable states via an agent-centric tree of asymptotically shortest paths.
However, the roadmap typically contains thousands of paths, which
exceeds the reasoning capacity of an LLM-based decision-maker and
introduces extreme redundancy. Many of these paths are semantically
equivalent for high-level intent, differing only in low-level
geometric details such as local motion smoothness.
We therefore extract a sparser set of nodes (\textit{informative
viewpoints}) by applying a two-pass online filter.

\paragraph{Spatial thinning}
Since information gain computation requires computational heavy ray-casting,
we therefore apply an online Poisson-disc sampling first: a candidate is
accepted only if no existing viewpoint lies within a radius.
This produces a well-distributed, low-redundancy candidate set at
negligible cost.

\paragraph{Information-gain gating}
For each spatially accepted candidate~\(\theta \in \mathrm{SE}(3)\),
we define its information gain as the number of unknown, visible
voxels within its view frustum:
\begin{equation}
  \label{eq:information-gain}
  r\!\left(\theta\!\mid\!\mathcal{M}\right) = \left\lvert \left\{ v
    \in \mathcal{M} \cap \mathcal{F}\!\left(\theta\right) :
    \operatorname{I}_{\text{visible}}\!\left(v, \theta\right) \cdot
  \operatorname{I}_{\text{unknown}}\!\left(v\right) = 1 \right\} \right\rvert
\end{equation}
where \(\mathcal{M}\) is the UFOMap, \(\mathcal{F}(\theta)\) is the
view frustum of camera pose~\(\theta\),
\(\operatorname{I}_{\text{visible}}(v, \theta) = 1\) iff the ray
from~\(\theta\) to the center of voxel~\(v\) is unoccluded, and
\(\operatorname{I}_{\text{unknown}}(v) = 1\) iff~\(v\) is classified as unknown.
Informative viewpoints are selected from candidates whose information
gain exceeds a threshold~\(\tau_{\mathrm{IG}}\),
and they serve a similar role to frontiers in frontier-based
exploration~\cite{yamauchiFrontierbasedApproachAutonomous1997}.
This sampling-based method integrate naturally with our RT-RRT*
roadmap, as the informative viewpoints are generated as a byproduct of
tree expansion rather than requiring a separate grid search like
classical exploration frontiers.
Given \(N\) retained informative viewpoints, we can query the RT-RRT* for the
current shortest path to each by tracing parent nodes in the tree structure.

%% file: methodology/option/b.tex
\subsubsection{Euclidean Steiner Tree Optimizer}
\begin{figure}[htbp]
  \centering
  \includegraphics[width=.43\linewidth]{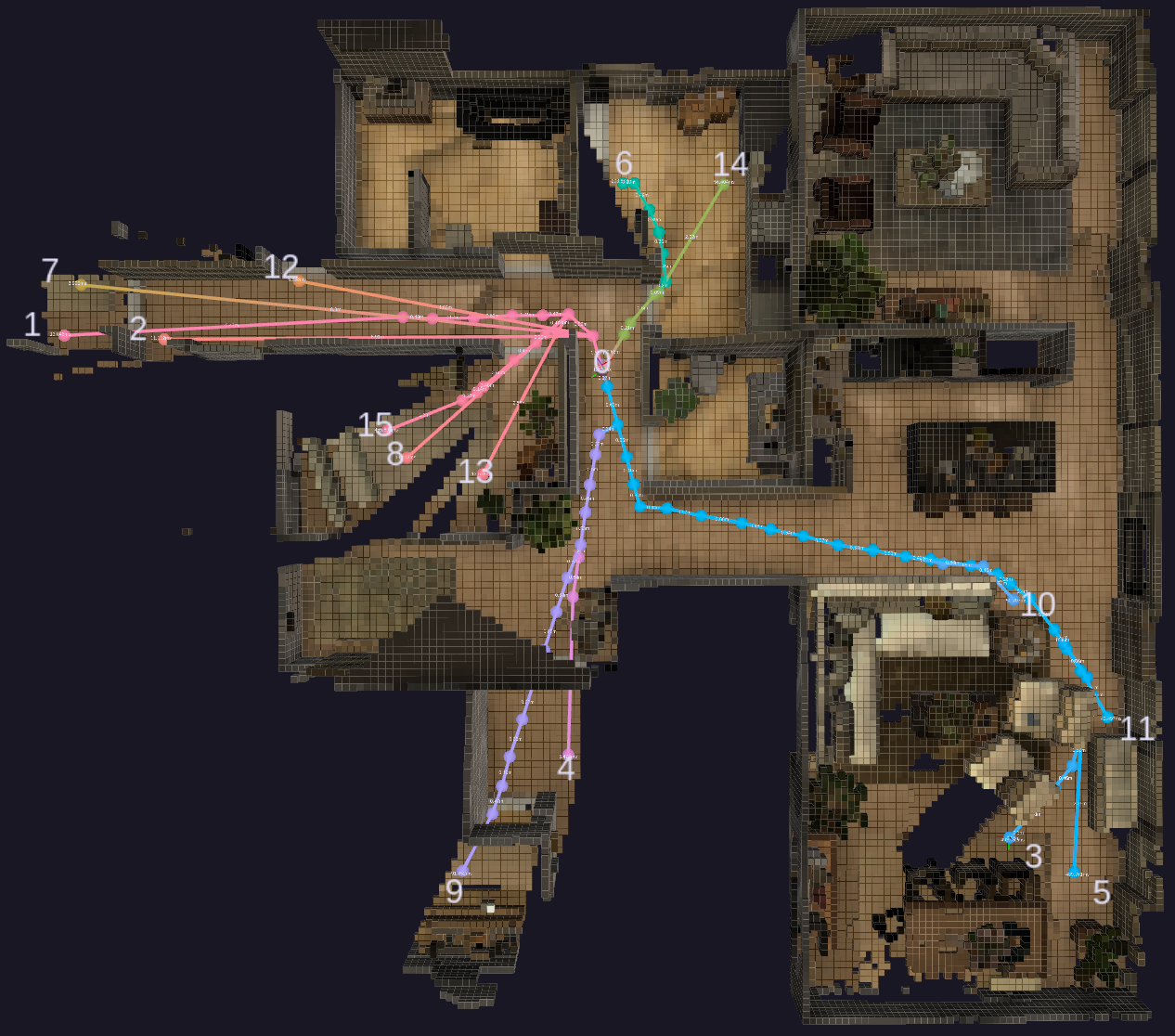}
  \;
  \includegraphics[width=.43\linewidth]{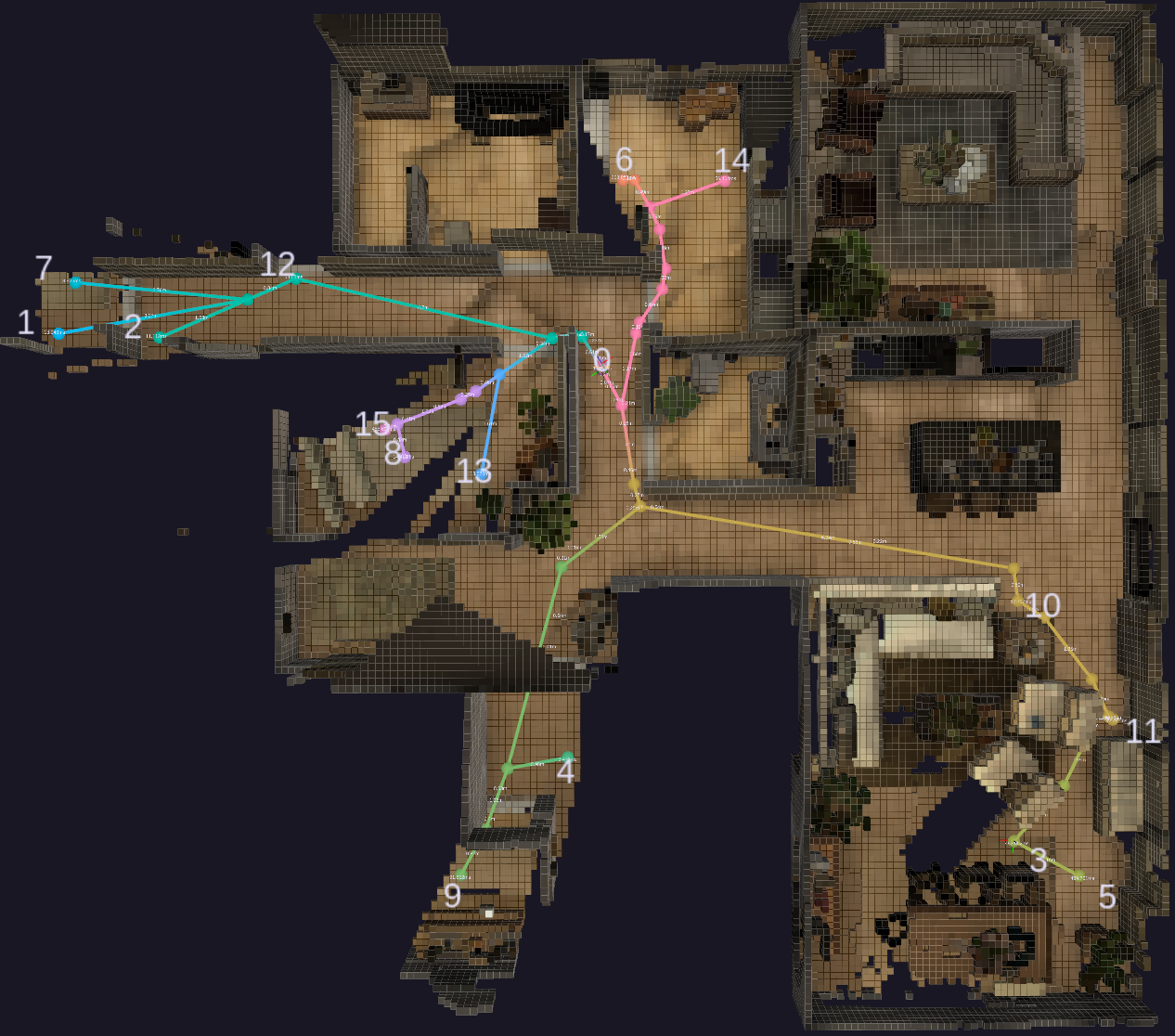}
  \caption{The RT-RRT* subtree connecting current agent (indexed by
    0) to all informative viewpoints indexed from 1 to 15 (left) versus
    the optimized Steiner tree (right). Independent per-path
    optimization produces redundant edges, while the OAESMT formulation
    merges shared segments and surfaces decision junctions, reducing
  total path length from \(85 \mathrm{m}\) to \(47 \mathrm{m}\).}
  \label{fig:pre-post-esto}
\end{figure}

Though RT-RRT* provides a tree of collision-free paths to informative
viewpoints,
because each path is optimized independently, unnecessary redundant
edges exist for navigation decision-making (\cref{fig:pre-post-esto}, left).
Our key insight is that sacrificing per-path optimality for a
globally shorter tree also reveals a hierarchical decision structure:
paths toward nearby terminals save total cost by sharing a common
segment and splitting only where the detour penalty exceeds the
sharing benefit, producing a trunk-to-branch-to-leaf topology,
visualized in ``Euclidean Steiner Tree Optimizer'' in \cref{fig:framework}.

Formally, let the agent pose \(r \in \mathcal{F}\) be the root and
the informative viewpoints \(V_T = \{v_1, \dots, v_N\} \subset
\mathcal{F}\) be terminals, where \(\mathcal{F} \subseteq
\mathbb{R}^2\) is obstacle-free space.
We seek a tree \(T = (V, E)\) embedded in~\(\mathcal{F}\) with
\(\{r\} \cup V_T \subseteq V\) that minimizes total edge cost \(C(T)
= \sum_{e \in E} \lVert e \rVert_2\).
This is an instance of the \emph{obstacle-avoiding Euclidean Steiner
Minimum Tree} (OAESMT~\cite{Zachariasen1999}) problem, which is
NP-hard even without obstacles:
the agent pose and informative viewpoints are the \emph{Steiner terminals},
the nodes any feasible tree must connect, while the solver may
introduce auxiliary \emph{Steiner points} to reduce total edge cost
by merging nearby path segments.
\par We approximate the solution in \cref{alg:esto}. A critical
observation is that the RT-RRT* subtree connecting root to all
terminals is already a feasible Steiner tree, just not at minimum cost.
We can therefore perform iterative local improvement on this warm start.
In each iteration, we generate candidate Steiner points by computing
the geometric median (Weiszfeld's
algorithm~\cite{weiszfeld1937point}) of each branching node and its
children via post-order traversal, add collision-free candidates to
the node set, compute a Minimum Spanning Tree over all nodes using
Kruskal's algorithm~\cite{kruskal1956shortest} with edges restricted
to collision-free line-of-sight segments, and prune non-terminal leaves.
The output tree is fed back as input until total cost no longer
decreases, converging to a local minimum.
Each iteration yields a valid tree, so it can act on the current best
result while refinement continues.
In practice, we refines the tree in the background and serves the
best-so-far result on demand; when the underlying planning problem
changes (\eg, new viewpoints appear), the solver restarts from the
updated warm start.

Viewpoints may become intermediate nodes when doing so reduces total
cost; the only invariant is that every viewpoint remains connected.
The resulting navigation decision tree surfaces decision points at
topological junctions such as corridor forks and room entrances
(\cref{fig:pre-post-esto}, right).
The LLM planner selects the next-best subtree at topological
junctions and commits to it in a receding-horizon fashion
(Sec.~\ref{sec:policy}).

\begin{algorithm}[htbp]
  \caption{Navigation Decision Tree Generation}\label{alg:esto}
  \KwInput{RT-RRT* tree $\mathcal{T}_{\mathrm{rrt}}$, agent pose $r$,
  informative viewpoints $V_T$, collision checker $\mathcal{CC}$}
  \KwOutput{Navigation decision tree $T^*$}
  $T \gets \textsc{ExtractSubtree}(\mathcal{T}_{\mathrm{rrt}},\, r,\,
  V_T)$;\quad $C^* \gets C(T)$\;
  \Repeat{$C(T') \geq C^*$}{
    $V \gets \mathrm{Nodes}(T)$;\quad $V_T \gets \{v \in V : v \text{
    is root or goal}\}$\;
    \ForEach{branching node $v \in T$ \textnormal{(post-order)}}{
      $\mathbf{s} \gets \textsc{WeiszfeldMedian}(\{v\} \cup
      \mathrm{Children}(v))$\;
      \lIf{$\exists\, \mathbf{s}^* \!\in\!
        \mathcal{T}_{\mathrm{rrt}}$ s.t.\ $\lVert \mathbf{s} {-}
      \mathbf{s}^* \rVert {<} \delta$}{$\mathbf{s} \gets \mathbf{s}^*$}
      \lIf{$\textsc{CollisionFree}(\mathbf{s})$}{$V \gets V \cup
      \{\mathbf{s}\}$}
    }
    $E \gets \{(u,v) : u,v \in V,\; \textsc{CollisionFree}(u, v)\}$\;
    $T' \gets \textsc{PruneLeaves}(\textsc{Kruskal}(V, E),\; V_T)$\;
    \lIf{$C(T') < C^*$}{$T^* \gets T'$;\; $C^* \gets C(T')$;\; $T \gets T'$}
  }
  \Return{$T^*$}
\end{algorithm}

%% file: methodology/policy/a.tex
\subsubsection{Tree Narration}
\par Given the geometric map, the semantic entity map
(Sec.~\ref{sec:belief}), and a constructed agent-centric navigation
decision tree, our goal is to convert this tree structure into a
linguistic representation that an LLM can reason about and plan.
We achieve this through a three-stage pipeline: (1) per-edge
annotation, (2) tree-level assembly, and (3) subtree captioning.

\paragraph{Per-Edge Annotation}
We decompose the navigation decision tree into individual edges and
annotate each edge in parallel.
Along each edge, we simulate a forward-facing virtual camera
traversing from the parent node to the child node, querying the
geometric and semantic maps for visible content.
At designated informative viewpoints (identified during tree
construction), the virtual camera switches to a panoramic field of
view, allowing a full scan of the surroundings at key decision points.
Each edge produces a structured description with two categories:
\textit{Known:} for each visible semantic entity, we log its
unique identifier, label, caption, and the sighting distance
along the edge from the parent node (\eg, ``pet carrier @ 0.5\,m,
sofa @ 2.0\,m'').
\textit{Unknown:} we cluster visible unknown voxels via
DBSCAN, computing each cluster's volume and its centroid distance
to the nearest known entity (\eg, ``3.2\,m\textsuperscript{3}
unknown region near the sofa'').

\paragraph{Tree-Level Assembly}
We assemble per-edge annotations by top-down traversal from
root to leaves.
Since each entity carries a unique identifier, we track sighting
distances across parent-child edges to detect spatial trends. If an
entity's sighting distance decreases along a root-to-leaf path, the
assembly marks this subtree as \textit{approaching} that entity.

\paragraph{Subtree Captioning}
A summarizer LLM converts the assembled structured annotations into
concise natural language options, one per root-level subtree.
The summarizer is encouraged to produce condensed, semantically rich
descriptions: individual object lists are abstracted into
region-level or room-level characterizations.
For example, a subtree containing \{chairs, dining table\} may be
captioned as ``dining area,'' while a subtree observing \{coffee
table, sofa, pet carrier, cat litter bin, stuffed animals\} becomes
``cluttered cozy living room with a pet-care area.''
Language descriptions of subtree-grounded options are fed to a
decision-making LLM later.

%% file: methodology/policy/b.tex
\subsubsection{Receding-Horizon LLM Planning}
\par The option-space pipeline (Sec.~\ref{sec:option}) runs in a
model-predictive-control (MPC) style loop: the RT-RRT* re-roots to
the agent's current pose, and the informative-viewpoint sampler with
the Steiner tree optimizer refresh the navigation
decision tree at \(10\,\mathrm{Hz}\).
The LLM deliberation, by contrast, requires on the order of seconds
per invocation.
We therefore decouple the system into two layers: a \emph{fast
reactive layer} that maintains the option space continuously, and an
\emph{event-triggered deliberative layer} that invokes the LLM only
at decision-relevant moments.
Because the reactive layer updates the tree regardless of the agent's
current commitment, new map changes are immediately reflected in the
option space; only when such changes are
decision-relevant does the deliberative layer re-engage.

\paragraph{Commitment Model}
Upon selection, the agent follows the chosen path through
non-branching nodes without re-querying the LLM.
The execution granularity of the receding-horizon loop is therefore
not a fixed metric segment but an adaptive path between consecutive
decision events, shaped by the topological structure of the environment.
This avoids redundant LLM invocations along corridors and other
topologically simple segments where no meaningful alternative exists.

\paragraph{Re-Invocation Triggers}
The LLM is re-invoked when the agent's current position is a
branching node in the live decision tree, i.e., the root of the
continuously updated tree has multiple child subtrees.
Because the tree updates in real time, this condition captures both
physical arrival at a pre-existing junction and the dynamic emergence
of a new branch at the agent's current position.
Re-invocation also occurs when a decision-relevant structural change
is detected within the committed subtree: a committed node is removed
(its viewpoint is no longer informative) or the subtree's
parent-child topology is rewired by the Steiner tree optimizer.
Changes outside the committed subtree are deferred until the agent
reaches a branching node, avoiding unnecessary re-invocations from
distant map updates.

\paragraph{Chain-of-Thought Scoring}
The LLM receives the target object category and subtree captions,
along with an special ``none of the above'' option that
permits abstention.
We prompt the reasoning LLM to analyze each option by estimating about the
likelihood of finding the target based on commonsense semantic co-occurrence and
spatial layout priors at first, then assigns a numeric score (0--100)
to each option.
For instance, when searching for a bed, the model might reason:
\textit{``Option~A describes a living room, which rarely contains
  beds. Option~B leads to a hallway near a large unexplored region,
which could plausibly lead to a bedroom.''}

\paragraph{Geometric Fallback}
When the LLM selects ``none of the above,'' the semantic signal is
insufficient for informed decision-making.
This occurs in two common scenarios: early in exploration when the
map contains few recognized entities, and in monolithic scenes where
visually similar paths (\eg, multiple featureless hallways at a
junction) offer no discriminative semantic cues.
In such cases, the agent falls back to a purely geometric strategy,
navigating to the nearest informative viewpoint.
Upon arrival, the full pipeline immediately reruns with the enriched
belief state.

%% file: experiments/experiments.tex
\subsection{Experimental Setup}

\paragraph{Datasets and configurations}
We evaluate our methodology in the Habitat simulator using three
photorealistic indoor datasets:
Gibson~\cite{xiaGibsonEnvRealWorld2018,chaplotObjectGoalNavigation2020},
HM3D~\cite{ramakrishnanHabitatmatterport3DDataset2021}, and
HSSD~\cite{khannaHabitatSyntheticScenes2024}.
We follow the conventional setup of the Habitat Navigation Challenge
2023~\cite{habitatchallenge2023} benchmark,
where the embodied agent is modeled as a grounded robot, equipped
with a head-mounted, forward-facing RGB-D camera that provides \(640
\times 480\) resolution observations at a \(79\degree\) horizontal
field-of-view (HFOV) with a depth sensing range of \([0.5 \mathrm{m},
5.0 \mathrm{m}]\).
The agent's action space consists of six discrete commands:
\texttt{MOVE\_FORWARD} (\(0.25 \mathrm{m}\)), \texttt{TURN\_LEFT}
($30\degree$), \texttt{TURN\_RIGHT} ($30\degree$), \texttt{LOOK\_UP}
($30\degree$), \texttt{LOOK\_DOWN} ($30\degree$), and \texttt{STOP}.

\paragraph{Evaluation metrics}
We report Success Rate (SR) and Success weighted by Path Length (SPL)
to evaluate ObjectNav performance.
SR measures the proportion of episodes successfully completed within
a strict budget of 500 action steps.
SPL further assesses navigation efficiency by weighting successful
episodes against the optimal path length.

\paragraph{Implementation details}
For off-the-shelf foundation models, we use a 2B-parameter, 8-bit
quantized variant of the instruction-tuned
Qwen3-VL~\cite{baiQwen3VLTechnicalReport2025} as the SVLM for both
semantic perception (\cref{subsubsection:semantic-mapping}) and
LLM-based summarization and reasoning (\cref{sec:policy}). We also use
\cite{Baccianella_JSON_Repair_2025} to mitigate JSON quality from the SVLM.
We use YOLO-World~\cite{chengYOLOWorldRealTimeOpenVocabulary2024} for
open-vocabulary object detection and
EdgeTAM~\cite{zhouedgetamondevicetrack2025} for instance segmentation
with box prompts.
All experiments are executed on a desktop computer with an NVIDIA
GeForce RTX 4080 GPU, an Intel Core i7-14700K CPU, and 32\,GB RAM.
We set the UFOMap voxel size to \(5\,\mathrm{cm}\), the
information-gain threshold to \(\tau_{IG} = 10\), and the
Poisson-disc radius for informative viewpoint spacing to \(r =
1.25\,\mathrm{m}\).
We run DBSCAN at \(1\,\mathrm{Hz}\) and use an IoU threshold of
\(\tau_{IoU} = 0.5\) to track entities across snapshots. For
cross-label entity merging, we use an IoS threshold of \(\tau_{IoS} =
0.3\) and a cosine similarity threshold of \(\tau_{sim} = 0.8\).

\subsection{Baseline Methods}

We compare REST against recent zero-shot ObjectNav methods.
VLFM~\cite{yokoyamaVLFMVisionLanguageFrontier2024} scores frontiers via VLM cosine similarity with the goal description.
ApexNAV~\cite{zhangApexNAVAdaptiveExploration2025} adaptively switches between exploration and exploitation frontiers with target-centric semantic fusion to denoise 2D semantics.
GAMap~\cite{fangGAMapZeroShotObject2024} augments a 2D semantic map with goal-oriented affordance and geometry embeddings.
SG-Nav~\cite{yinSGNavOnline3D2024} constructs an online 3D scene graph for frontier selection to unleash LLM reasoning capabilities.
VoroNav~\cite{wuVoroNavVoronoibasedZeroshot2024} simplifies free space into a Voronoi graph for topological planning.
TriHelper~\cite{zhangTriHelperZeroShotObject2024} decomposes navigation into three specialized sub-policies for exploration, approach, and identification.
ImagineNav~\cite{zhaoImagineNavPromptingVisionlanguage2025} prompts a VLM to imagine unobserved regions in a mapless framework.
UniGoal~\cite{yinUniGoalUniversalZeroshot2025} extends SG-Nav to unify goal representations across navigation tasks.
PanoNav~\cite{jinPanoNavMaplessZeroShot2025} performs mapless navigation from panoramic VLM reasoning.

\input{experiments/benchmark-comparison}

\input{experiments/ablation-study}

%% file: experiments/benchmark-comparison.tex
\subsection{Benchmark Comparison}
Table~\ref{tab:main-results} reports SR and SPL across three datasets.
\begin{table}[htbp]
  \centering
  \caption{Comparison of zero-shot methods on Gibson, HM3Dv1, and
    HSSD ObjectNav validation splits. {\colorbox{color-first}{First}},
    {\colorbox{color-second}{second}}, and
  {\colorbox{color-third}{third}} results are highlighted.}
  \label{tab:main-results}
  \resizebox{.75\linewidth}{!}{
    \begin{tblr}{
        colspec={l*{6}{Q[c,m]}},
        cells={mode=math},            %
        column{1}={mode=text},        %
        row{1-2}={mode=text},         %
        hline{2}={2-3}{leftpos=-1,rightpos=-1,endpos},
        hline{2}={4-5}{leftpos=-1,rightpos=-1,endpos},
        hline{2}={6-7}{leftpos=-1,rightpos=-1,endpos},
      } %
\toprule
\SetCell[r=2]{l} Method
& \SetCell[r=1,c=2]{c} Gibson &
& \SetCell[c=2]{c} HM3Dv1 &
& \SetCell[r=1,c=2]{c} HSSD &
\\
& SR   & SPL  & SR   & SPL  & SR   & SPL  \\
\midrule
SG-Nav%
& -    & -    & 54.0 & 24.9 & -    & -    \\
GAMap%
& \SetCell{bg=color-first} 85.7 & \SetCell{bg=color-first} 55.5 & 53.1 & 26.0 & -    & -    \\
VoroNav%
& -    & -    & 42.0 & 26.0 & \SetCell{bg=color-third} 41.0 & \SetCell{bg=color-third} 23.2 \\
VLFM%
& 84.0 & \SetCell{bg=color-third} 52.2 & 52.5 & \SetCell{bg=color-third} 30.4 & -    & -    \\
TriHelper%
& \SetCell{bg=color-second} 85.2 & 43.1 & \SetCell{bg=color-third} 56.5 & 25.3 & -    & -    \\
ApexNAV%
& -    & -    & \SetCell{bg=color-first} 59.6 & \SetCell{bg=color-second} 33.0 & -    & -    \\
ImagineNav%
& -    & -    & 53.0 & 23.8 & \SetCell{bg=color-second} 51.0 & \SetCell{bg=color-second} 24.9 \\
UniGoal%
& -    & -    & 54.5 & 25.1 & -    & -    \\
PanoNav%
& -    & -    & 43.5 & 23.7 & -    & -    \\
\textbf{REST (Ours)}
& \SetCell{bg=color-third} 85.1 & \SetCell{bg=color-second} 53.5 & \SetCell{bg=color-second} 57.3 & \SetCell{bg=color-first} 33.4 & \SetCell{bg=color-first} 56.7 & \SetCell{bg=color-first} 29.1 \\
\bottomrule
      \end{tblr}
    }
  \end{table}
  On Gibson, REST achieves competitive SR and SPL matching the
  top-performing method.
  The compact layouts of Gibson scenes leave limited room for
  differentiation, while strong methods already approach saturation
  on this dataset.
  On HM3D, the larger and noisier real-world scans present a more
  discriminative benchmark.
  ApexNAV achieves the highest SR,
  while REST attains competitive SR with the highest SPL across all methods.
  We attribute the SR gap primarily to scanning artifacts in HM3D;
  for example, reflective surfaces (mirrors, glass) are captured as
  geometric holes that introduce phantom openings and inflate the 3D
  information gain calculation, repeatedly drawing the agent to poses
  where no real progress can be made and exhausting the 500-step budget.
  By contrast, 2D frontier methods such as ApexNAV are largely immune.
  This sensitivity to scan noise reduces SR, yet episodes that REST
  does complete benefit from our tree-structured and path-grounded
  planning, yielding the highest SPL.
  On HSSD, among the methods that report results, REST leads in both SR and SPL.
  HSSD scenes feature diverse layouts (garages, outdoor lounges,
  extended corridors) with sparser semantic cues than Gibson or HM3D,
  which degrades point-wise waypoint ranking: the highest-similarity
  score can attach to an irrelevant nearby object, trapping the agent
  in a greedy local minimum.
  REST's option space structures waypoints into a tree of paths,
  letting the LLM evaluate directions rather than individual
  destinations, a coarser granularity that remains discriminative
  even when semantics are sparse in the environment.
  Compared with VoroNav~\cite{wuVoroNavVoronoibasedZeroshot2024}, the
  closest graph-based baseline, REST improves SR by over \(15\)
  points on both HM3Dv1 and HSSD and raises SPL by \(7.4\) and
  \(5.9\) points, respectively.
  Since both reason over a navigational graph, this gap isolates the
  payoff of our belief-driven topology, decoupled from known
  geometry, over VoroNav's geometry-locked Voronoi skeleton.

%% file: experiments/ablation-study.tex
\subsection{Ablation Study}

\par We ablate REST's two core components on HSSD to isolate their individual contributions.
\begin{table}[htbp]
	\centering
	\caption{Ablation study on HSSD ObjectNav validation split.}
	\label{tab:ablation}
	\begin{tblr}{
		colspec={lQ[c,m]Q[c,m]},
		cells={mode=math},
		column{1}={mode=text},
		row{1}={mode=text},
		}
		\toprule
		Variant & SR   & SPL  \\
		\midrule
		REST (Full)
		        & 56.7 & 29.1 \\
		w/o LLM reasoning
		        & 29.1 & 18.7 \\
		w/o Steiner tree
		        & 53.1 & 25.3 \\
		\bottomrule
	\end{tblr}
\end{table}
Without LLM reasoning, the agent defaults to the nearest informative viewpoint and exhausts the 500-step budget more frequently, yielding lower SR and substantially lower SPL.
More notably, removing the Steiner tree optimizer while retaining LLM reasoning reveals a compounding degradation across the pipeline.
Without compaction, the raw RT-RRT* subtree branches early and shares shorter trunks, so edges that would have been merged into a single segment remain separate.
This inflation multiplies the tree narrator's workload: more edges must be ray-cast in parallel, increasing wall-clock latency per decision cycle.
Worse, the redundant edges produce repetitive tokens in the subtree captioning stage, degrading the performance of our deployed SVLM summarizer.
The resulting option descriptions presented to the decision-making LLM are longer and more redundant, obscuring the topological decision points (corridor forks, room entrances) that the Steiner tree explicitly surfaces.
Both SR and SPL decrease relative to the full model, confirming that compact decision trees are a more effective substrate for LLM spatial reasoning than individually optimized paths.

%% file: conclusion/conclusion.tex
\par We argued that the option space of a zero-shot ObjectNav agent should be a tree of paths rather than independently scored waypoints.
REST instantiates this idea with a Receding Horizon Explorative Steiner Tree whose branches are textualized via an open-vocabulary 3D semantic map and selected through hierarchical LLM reasoning.
Experiments on Gibson, HM3D, and HSSD show that REST consistently ranks among the top methods in success rate while achieving the best or second-best path efficiency, confirming a favorable efficiency-success balance.
These results suggest that rethinking the option space, the interface between perception and decision-making, is a complementary and underexplored axis for improving future hierarchical embodied AI agents.